\documentclass[final,times,sort&compress]{elsarticle}

\usepackage{lineno,hyperref}
\modulolinenumbers[5]

\usepackage{longtable}
\usepackage{tabulary}
\usepackage{graphicx}

\usepackage{subfig}
\usepackage [english]{babel}
\usepackage [autostyle, english = american]{csquotes}
\MakeOuterQuote{"}

\usepackage{listings}
\usepackage{enumerate} 
\usepackage[section]{placeins}
\usepackage{amsmath}

\usepackage{booktabs}
\usepackage{multirow}
\usepackage{float}

\usepackage{fancyhdr}
\usepackage{textcomp}

\usepackage{threeparttable}

\fancypagestyle{plain}{%
  \fancyhf{}
  \fancyfoot[C]{\iffloatpage{}{\thepage}}
  }
\journal{ 2020 IEEE  CONECCT conference }
 
\begin{document}

\begin{frontmatter}	

\title{A Hybrid Deep Learning Approach for Diagnosis of the Erythemato-Squamous Disease}
\tnotetext[mytitlenote]{Pre-review version of the paper accepted at the 2020 IEEE CONECCT Conference.\\
      978-1-7281-6828-9/20/\$31.00~\copyright~2020 IEEE}


%

\author[myu]{Sayan Putatunda \corref{cor1}}
\ead{sayanp@iima.ac.in}

\address[myu]{Indian Institute of Management Ahmedabad, India}


\begin{abstract}
The diagnosis of the Erythemato-squamous disease (ESD) is accepted as  a difficult problem in dermatology. ESD is a form of skin disease. It generally causes redness of the skin and also may cause loss of skin. They are generally due to genetic or environmental factors. ESD comprises six classes of skin conditions namely, pityriasis rubra pilaris, lichen planus, chronic dermatitis, psoriasis, seboreic dermatitis and pityriasis rosea. The automated diagnosis of ESD can help doctors and dermatologists in reducing the efforts from their end and in taking faster decisions for treatment. The literature is replete with works that used conventional machine learning methods for the diagnosis of ESD. However, there isn't much instances of application of Deep learning for the diagnosis of ESD.

In this paper, we propose a novel hybrid deep learning approach i.e. Derm2Vec for the diagnosis of the ESD. Derm2Vec is a hybrid deep learning model that consists of both Autoencoders and Deep Neural Networks. We also apply a conventional Deep Neural Network (DNN) for the classification of ESD. We apply both Derm2Vec and DNN  along with other traditional machine learning methods  on a real world dermatology dataset. The Derm2Vec method is found to be the best performer (when taking the prediction accuracy into account) followed by DNN and Extreme Gradient Boosting.The mean CV score of Derm2Vec, DNN and Extreme Gradient Boosting are $96.92\%$, $96.65\%$ and $95.80\%$ respectively.

\end{abstract}


\end{frontmatter}


\section{Introduction}
\label{intro}
Erythemato-squamous disease (ESD) is a form of skin disease. It generally causes redness of the skin and also may cause loss of skin. ESDs are generally due to genetic or environmental factors \citep{esd:2}.  ESD comprises six classes of skin conditions namely, pityriasis rubra pilaris, lichen planus, chronic dermatitis, psoriasis, seboreic dermatitis and pityriasis rosea. However, the diagnosis of ESD is accepted as a difficult problem in Dermatology. The reason why ESD is difficult to diagnose is due to the fact that these diseases share many clinical and histopathological attributes with erythema and scaling. Another reason is that one disease may show the symptoms of another disease at the initial stages \citep{esd:1}. Thus, a detailed observation skills and high experience are required from physicians to evaluate both clinical and histopathological features to correctly diagnose ESD \citep{esd:3}. So, the automated diagnosis of ESD can help doctors and dermatologists in reducing the efforts from their end and in taking faster decisions for treatment. 

In the literature, there are a quite a few instances of works that proposed various machine learning methods such as decision trees, support vector machines, artificial neural networks and more for automated detection of the type of Erythemato-squamous disease \citep{esd:3}. We will discuss in detail about these works in Section \ref{lit}. In recent years, with the rise in computing power and availability of cheap memory devices along with cloud computing, deep learning has been very successful in many fields such as natural language processing \citep{dl:7}, biomedicine \citep{dl:6}, computer vision \citep{dl:8} and more. 

The main contribution of this paper is in the development of a novel hybrid deep learning approach i.e. Derm2Vec for the diagnosis of the Erythemato-Squamous disease (ESD) that hasn't been reported in the literature earlier to the best of our knowledge. Derm2Vec is a hybrid deep learning approach that comprises both Autoencoders and Deep Neural Networks. Also, we find that there haven't been many works reported in the literature regarding the applications of deep neural networks for the classification of ESD. Although the literature is replete with works that used conventional machine learning methods (such as Random forests, artificial neural netwroks, Extreme Gradient Boosting, K-nearest neighbors, decision trees, support vector machines and more) for the diagnosis of ESD. In this paper, we apply both Derm2Vec and DNN (after tuning the hyperparameters) along with other conventional machine learning methods on a real world dermatology dataset. The Derm2Vec method is found to be the best performer when taking the prediction accuracy into account.

The rest of this paper is organized as follows. In Section \ref{lit}, we present a brief review of the literature.  In Section \ref{data}, we describe the dataset used in this paper. This is followed by Sections \ref{method} and \ref{exp} where we describe our proposed methodology and the experimental results respectively.  Finally, Section \ref{conclusion} concludes the paper.

\section{Related Work} \label{lit}
There have been many works in the medical informatics literature on the applications of machine learning and expert systems and how it complements physicians/practitioners in decision making. For example, Cruz \& Wishart \cite{lit:1} used machine learning for cancer prediction. Random forests was used for classification of the Alzheimer's disease \citep{lit:3}. Deep learning was used for the classification of self-care problems in children with physical disabilities \cite{arxiv:1}. There have been many applications of machine learning for glaucoma screening \citep{lit:4}, retinal hemorrhage detection \citep{lit:5}, lymphoma classification \citep{lit:6} and many more.

Automated classification of the type of Erythemato-squamous disease using machine learning and expert systems is reported in the literature. The first such work is that of Demiroz et al. \cite{esd:1} where the authors developed a new classifier called "Voting feature intervals-5" for the differential diagnosis of ESD. Guvenir \& Emeksiz \cite{esd:4} used three classification algorithms namely, Voting feature intervals-5, Na{\"i}ve Bayes and nearest neighbor classification for diagnosis of the type of ESD. Ubeyli \cite{esd:5} used multi layer perceptron neural networks and Xie \& Wang \cite{esd:6} used support vector machines for the classification of ESD. Even tree based methods such as CHAID decision trees \citep{esd:2} and ensemble of decision trees \citep{esd:3} have been used for analysis and diagnosis of ESD. Nanni \cite{esd:9} used an ensemble of support vector machines on random subspace and Menai \cite{esd:11} applied random forests for the diagnosis of ESD.

Some of the other interesting methods reported in the literature for the diagnosis of ESD are fuzzy classification \citep{esd:8}, neuro-fuzzy inference systems \citep{data:1}, k-means clustering \citep{esd:7}, boosting \citep{esd:12} and genetic programming \citep{esd:10}. We find that none of the past works reported in the literature have used Deep learning to the best of our knowledge. In this paper, we use Deep neural networks for the differential diagnosis of ESD and propose a novel hybrid deep learning method i.e. Derm2Vec (see Section \ref{derm2vec}).

\section{Dermatology Data} \label{data}
In this paper, we use the dermatology dataset that was first used by Ubeyli \& Guler \cite{data:1} where the aim was to determine the type of ESD. This dataset is publicly available in the UCI machine learning repository \citep{uci:1}. The dataset contains $33$ attributes/predictor variables where $12$ are for clinical attributes (namely, (a) erythema, (b) scaling, (c) definite borders, (d) itching, (e) koebner phenomenon, (f) polygonal papules, (g) follicular papules, (h) oral mucosal involvement, (i) knee and elbow involvement, (j) scalp involvement, (k) family history and (l) Age). 

The remaining $21$ features are for the histopathological attributes (namely, (a) melanin incontinence, (b) eosinophils in the infiltrate, (c) PNL infiltrate, (d) fibrosis of the papillary dermis, (e) exocytosis, (f) acanthosis, (g) hyperkeratosis, (h) parakeratosis, (i) clubbing of the rete ridges, (j) elongation of the rete ridges, (k) thinning of the suprapapillary epidermis, (l) spongiform pustule, (m) munro microabcess, (n) focal hypergranulosis, (o) disappearance of the granular layer, (p) vacuolisation and damage of basal layer, (q) spongiosis, saw-tooth appearance of retes, (r) follicular horn plug, (s) perifollicular parakeratosis, (t) inflammatory monoluclear inflitrate and (u) band-like infiltrate). The total number of features after performing one-hot encoding for the categorical variables becomes $129$. 

The total number of observations in the dataset are $366$. However, there are around $8$ missing values for the "Age" variable in the dataset and we won't be considering these observations in our analysis. So, finally the dataset has $358$ instances after removing missing values. The target variables had $6$ classes and the number of instances in each are- (a) psoriasis- $111$, (b) seboreic dermatitis- $60$, (c) lichen planus- $71$, (d) pityriasis rosea- $48$, (e) chronic dermatitis- $48$ and (f) pityriasis rubra pilaris- $20$.

\section{Methodology} \label{method}
Sections \ref{ann} and \ref{dl} give a brief overview of artifical neural networks and deep learning respectively. In Section \ref{derm2vec}, we discuss our proposed method i.e. Derm2Vec and Section \ref{others} metions the different machine learning methods that we will use for comparison of performance in this paper.

\subsection{Artificial Neural Networks} \label{ann}
To understand Deep learning (discussed in Section \ref{dl}), first we need to understand Artifical Neural Networks (ANNs). The origin of ANNs can be traced to the study of information processing in life sciences \citep{ann:6, ann:3, ann:4}. McCulloch \& Pitts \cite{ann:6} worked on developing nets of simple logical operators to model biological systems. Rosenblatt \cite{ann:8} introduced the concept of "perceptron" that is a biologically inspired learning algorithm. Neural networks are used for various statistical modeling and data analysis tasks and it is seen as an alternative to non-linear regression \citep{ann:7}.

In this paper, we focus on the "feed-forward neural networks". Bishop \cite{ann:1} describes a two layered feed-forward neural network architecture consisting of an input layer that is followed by a hidden layer (that consists of hidden nodes) and finally, an output layer. The hidden nodes are like processing units that contains activation functions. Some of the commonly used activation functions are Sigmoid, ReLu, tanh and more \citep{sayan:1}. A feed forward neural networks performs layered computations where the hidden unit activations are computed using the input layer and then the output is calculated using the hidden unit activations \citep{ann:2}. Please refer to Bishop \cite{ann:1} for more details on the modus-operandi of the feed-forward neural networks.

Nowadays, artificial neural networks are widely used in various applications such as weather forecasting \citep{ann:10}, clinical medicine \citep{ann:11}, Forex prediction \citep{ann:12}, Location/Travel time prediction for GPS Taxis \citep{sayan:3, sayan:2, thesis:sayan} and more. Please see Bishop \cite{ann:9} for more details on artificial neural networks.

\subsection{Deep learning: Deep Neural Networks and Autoencoders} \label{dl}
Goodfellow et al. \cite{dl:1} describes deep learning as a subset of machine learning and as a form of "representation learning". Here the focus is on using the raw data to extract high level features. Chollet \cite{dl:5} describes deep learning as learning from successive layers, each layer being some meaningful representation. In the recent years, deep learning has tasted success in various applications such as natural language processing \citep{dl:7}, biomedicine \citep{dl:6}, computer vision \citep{dl:8} and more. There are different kinds of deep learning architectures such as Recurrent neural networks and Convolutional neural networks (see Goodfellow et al. \cite{dl:1} for more details on different deep learning techniques). However, in this paper we will focus on Deep Neural Networks (DNNs) and Autoencoders.

Deep neural networks (DNNs) originated from Artificial neural networks (see Section \ref{ann}). ANNs with many hidden layers is known as DNNs \citep{dl:6}. These number of hidden layers determine the "depth" of a DNN \citep{dl:5}. An Autoencoder is a type of deep learning method where the input and the output are same. It is classified as self-supervised learning method by  Chollet \cite{dl:5}. An Autoencoder consists of two functions namely, (a) Encoder function- here the raw input data is converted into representations  and (b) Decoder function- here the representations from the encoder layer are converted back to the input data. The goal of an autoencoder is to preserve as much information as possible and also add new representations on top of the raw input data \citep{dl:1}. Some of the great applications of Autoencoders include dimensionality reduction \citep{auto:1}, cyber-emphatic design \citep{dl:9}, molecular design \citep{dl:10} and many more.

\subsection{Proposed Method: Derm2Vec} \label{derm2vec}
In this paper, first we apply a conventional DNN on the dataset for the prediction of the Erythemato-Squamous disease. The usage of DNN has't been reported in the literature (see Section \ref{lit}) for the diagnosis of ESD. Although ANNs have been used earlier as mentioned in Section \ref{lit}.

\begin{figure}[!htp]
 \centering
  \includegraphics[width=0.75\textwidth]{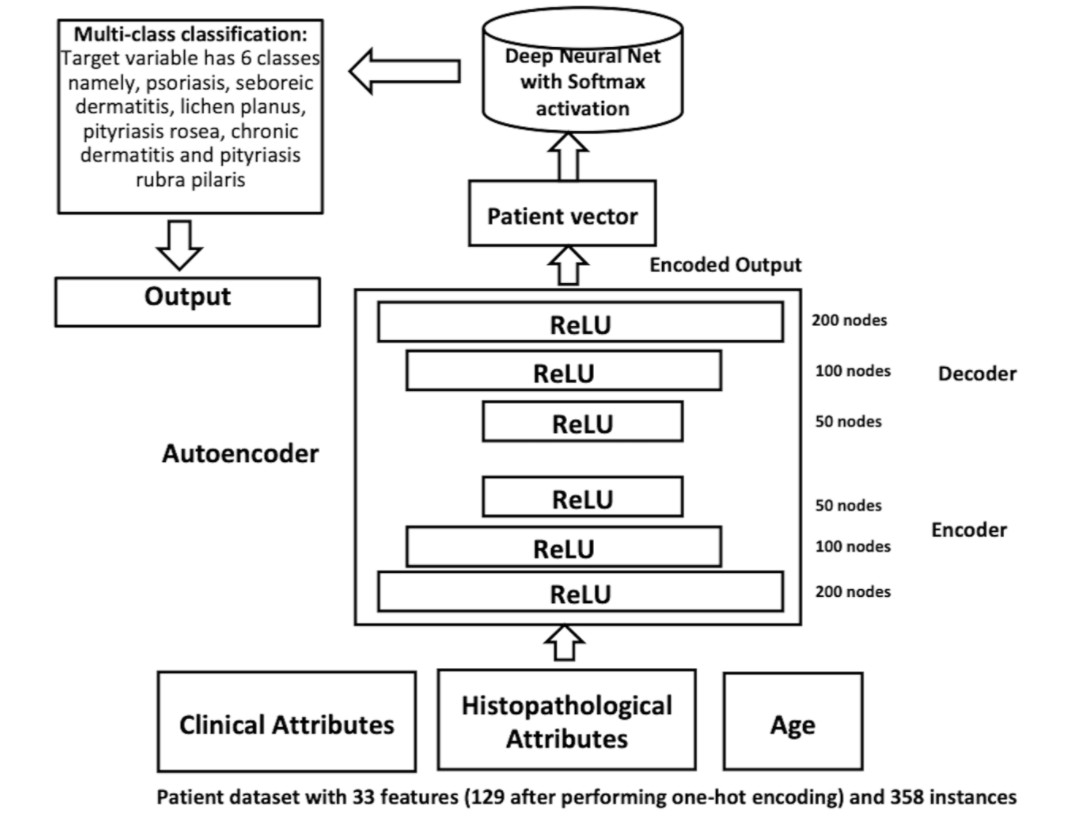}
  \caption{The Derm2Vec method}
  \label{fig:boat1}
\end{figure}

In this paper, we propose a novel hybrid deep learning approach that is a two-step modeling approach comprising an autoencoders and a DNN for multi-class classification of type of ESD. This hasn't been reported (to the best of our knowledge) in the dermatology informatics literature. We will refer to our proposed method as"Derm2Vec". Figure \ref{fig:boat1} shows the modus-operandi of the Derm2Vec method. 

We can see in Figure \ref{fig:boat1} that high dimensional the input data (i.e. $129$ features related to clinical and Histopathological attributes along with Age) is passed through an Autoencoder that comprises three encoder and three decoder layers containing $50$, $100$ and $200$ nodes in each. The different encoding dimensions used by the encoder are $4$, $8$, $16$, $32$, $40$, $48$ and more (see Section \ref{classi}). The values from the innermost encoder layer is taken i.e. the Encoded output that represents a dense patient vector. We then apply a DNN (comprising a single hidden layer with $100$ nodes or two hidden layers with $100$ nodes in each) on the patient vector to get the predicted output. Since, the target variable contains $6$ classes i.e. this is a multi-class classification problem, so we use the "Softmax" activation function in the output layer of the DNN.

\subsection{Other Methods} \label{others}
In this paper, we will compare the performance of both our proposed method i.e. Derm2Vec and a conventional DNN method with other conventional machine learning techniques that have been used for the prediction of the Erythemato-Squamous disease in the literature as discussed in Section \ref{lit}. Some of the techniques we will use in this paper for comparison are Decision trees \citep{dt:1,
stat:5}, Artificial Neural Networks (see Section \ref{ann}), ensemble learning methods such as Extreme Gradient Boosting \citep{xgb:1} and Random Forests \citep{stat:6}, K nearest neighbors \citep{knn:1}, Support Vector Classification \citep{svc:1} and finally, the Gaussian Na{\"i}ve Bayes \citep{nb:1}.

Please refer to \cite{ann:1}, \cite{stat:1}, \cite{stat:2} and \cite{sayan:1} for a detailed account on the various machine learning methods mentioned above.

\section{Experimental Results} \label{exp}
Our goal is to predict the type of Erythemato-Squamous disease using our proposed method i.e. Derm2Vec along with a Deep Neural Network on the dermatology dataset described in Section \ref{data}. This is a multi-class classification problem as the target variable has $6$ classes. We will compare the performance of the above mentioned method with other conventional machine learning techniques that have been used in the literature for the prediction of ESD (see Section \ref{lit}) and also mentioned in Section \ref{others} such as Extreme Gradient Boosting (XGBoost), Artificial Neural Network (ANN), Random Forest (RF), Decision Tree (DT), Na{\"i}ve Bayes (NB), K-nearest neighbors (KNN) and Support Vector Classification (SVC).

All the experiments are conducted in a system with mac OSX,  $16$ GB RAM and Intel core i7 processor. The data analysis and model development were done in Python \citep{py:1}. We use the scikit-learn \citep{sci:1}, Tensorflow \citep{py:3} and Keras \citep{py:2}  libraries for implementing the various machine learning and deep learning techniques used in this paper.

\subsection{Evaluation Metrics} \label{eval}
We apply the above mentioned methods on the dermatology dataset and perform k-fold cross-validation \citep{eval:1}. In this paper, we perform 10-fold cross-validation i.e. the dataset is partitioned into $10$ equal sets or folds and then $10$ subsequent iterations are performed where $9$ folds are used for model training and $1$ fold is withheld for validation. We use the Mean cross-validation score i.e. the Mean CV score as a evaluation metric for our experiments in this paper (see Section \ref{classi}). The Mean CV score is the average of the accuracy scores obtained in each iterations while performing the 10 fold cross-validation. 

A higher value of mean CV score indicates better performance of the method. It is reported in terms of percentage ($\%$) in this paper.

\subsection{Results: Classification of the Erythemato-Squamous Disease} \label{classi}
Table \ref{tab1} describes the mean CV scores when we apply Deep Neural Network (DNN) with different hyperparamters on the dataset and perform 10-fold cross validation. We run multiple iterations of DNN with different number of hidden layers such as $1$, $2$ or $3$ containing different hidden nodes such as $100$, $200$ and $300$. Since this is in a multi-class classification setting, so the output layer of the DNN will have a Softmax activation function. We also use "Dropouts" that ensure that the deep neural network model doesn't overfits \citep{dl:11}. Here, $Dropout=0.5$ means that $50\%$ of the units are dropped randomly during training. We find that the highest mean CV score for DNN (that consists of $1$ hidden layer, $100$ hidden nodes and $dropout=0.5$) is $96.65\%$.

\begin{table}[!htp]
\centering
\caption{Mean CV score for DNN with different hyperparameters}
\label{tab1}
 \scalebox{0.8}{
\begin{tabular}{ccccc}
\hline
\multirow{2}{*}{Sl no.} & \multicolumn{3}{c}{DNN hyperparameters}                                                                                                           & \multirow{2}{*}{\begin{tabular}[c]{@{}c@{}}Mean CV \\ score (\%)\end{tabular}} \\\cmidrule(l){2-4}
                        & \begin{tabular}[c]{@{}c@{}}No. of hidden\\  layers\end{tabular} & \begin{tabular}[c]{@{}c@{}}No. of hidden\\  nodes\end{tabular} & Dropouts  &                                                                                \\ \hline
1                       & 2                                                               & (100, 100)                                                     & No                        & 96.1                                                                           \\ \hline
2                       & 1                                                               & 100                                                            & No                        & 95.8                                                                           \\ \hline
3                       & 2                                                               & (100, 100)                                                     & Yes (0.5)                 & 96.37                                                                          \\ \hline
4                       & 1                                                               & 100                                                            & Yes (0.5)                 & 96.65                                                                          \\ \hline
5                       & 3                                                               & (100, 100, 100)                                                & No                        & 96.37                                                                          \\ \hline
6                       & 3                                                               & (100, 100, 100)                                                & Yes (0.5)                 & 96.08                                                                          \\ \hline
7                       & 1                                                               & 200                                                            & No                        & 95.8                                                                           \\ \hline
8                       & 1                                                               & 200                                                            & Yes (0.5)                 & 96.09                                                                          \\ \hline
9                       & 1                                                               & 300                                                            & No                        & 95.81                                                                          \\ \hline
10                      & 1                                                               & 300                                                            & Yes (0.5)                 & 95.25                                 \\     \hline                                          
\end{tabular}}
\end{table}

We now apply the Derm2Vec method on the dermatology dataset with different hyperparameters and run multiple iterations as described in Table \ref{tab2}. Here we tune the hyperparameters of the Autoencoder and the DNN. For the Autoencoder, the encoder and decoder comprises $3$ layers each with $50$, $100$ and $200$ nodes as shown in Figure \ref{fig:boat1}. The only hyperparameter we tune is the encoding dimensions i.e. we compress the high dimensional dataset (i.e. containing $129$ features) into a low dimensional space. We vary the encoding dimensions from $4$, $8$, $16$, $32$, $40$, $48$, \ldots, upto $88$ as described in Table \ref{tab2}. We also tune the hyperparameters of the subsequent DNN of Derm2Vec i.e. Dropouts, the number of hidden layers and the number of hidden nodes. We find that the highest mean CV score is $96.92\%$ that is higher than what we got for DNN in Table \ref{tab1}. Thus, Derm2Vec performs better than DNN. In fact from Tables \ref{tab1} and {tab2} we can clearly see that the best performing configuration of Derm2Vec i.e. DNN  with $1$ hidden layer, $100$ hidden nodes and $dropout=0.5$ when complemented with an Autoencoder with encoding dimension of $32$ perform better than the stand alone DNN with similar configuration (i.e. $1$ hidden layer, $100$ hidden nodes and $dropout=0.5$). This shows that the proposed hybrid deep neural network approach i.e. Derm2Vec is a better performer (when taking the prediction accuracy into consideration) than a conventional deep neural network.

\begin{table}[!htp]
\centering
\caption{Mean CV score for Derm2Vec with different hyperparameters}
\label{tab2}
 \scalebox{0.75}{
\begin{tabular}{cccccc}
\hline
\multirow{2}{*}{Sl no.} & \multirow{2}{*}{\begin{tabular}[c]{@{}c@{}}Autoencoder parameter-\\  Encoding dimensions\end{tabular}} & \multicolumn{3}{c}{DNN hyperparameters}                                                                                                           & \multirow{2}{*}{\begin{tabular}[c]{@{}c@{}}Mean CV \\ score (in \%)\end{tabular}} \\ \cmidrule(l){3-5}
                        &                                                                                                        & \begin{tabular}[c]{@{}c@{}}No. of hidden \\ layers\end{tabular} & \begin{tabular}[c]{@{}c@{}}No. of hidden\\  nodes\end{tabular} & Dropouts  &                                                                                   \\ \hline
1                       & 4                                                                                                      & 1                                                               & 100                                                            & Yes (0.5) & 95.52                                                                             \\\hline
2                       & 8                                                                                                      & 1                                                               & 100                                                            & Yes (0.5) & 96.37                                                                             \\\hline
3                       & 16                                                                                                     & 1                                                               & 100                                                            & Yes (0.5) & 95.79                                                                             \\\hline
4                       & 24                                                                                                     & 1                                                               & 100                                                            & Yes (0.5) & 96.35                                                                             \\\hline
5                       & 32                                                                                                     & 1                                                               & 100                                                            & Yes (0.5) & 96.92                                                                             \\\hline
6                       & 40                                                                                                     & 1                                                               & 100                                                            & Yes (0.5) & 94.96                                                                             \\\hline
7                       & 48                                                                                                     & 1                                                               & 100                                                            & Yes (0.5) & 96.37                                                                             \\\hline
8                       & 32                                                                                                     & 2                                                               & 100                                                            & No        & 94.97                                                                             \\\hline
9                       & 32                                                                                                     & 2                                                               & 100                                                            & Yes (0.5) & 96.92                                                                             \\\hline
10                      & 32                                                                                                     & 1                                                               & 100                                                            & Yes (0.5) & 96.1                                                                              \\\hline
11                      & 56                                                                                                     & 1                                                               & 100                                                            & Yes (0.5) & 96.94                                                                             \\\hline
12                      & 64                                                                                                     & 1                                                               & 100                                                            & Yes (0.5) & 95.52                                                                             \\\hline
13                      & 72                                                                                                     & 1                                                               & 100                                                            & Yes (0.5) & 95.53                                                                             \\\hline
14                      & 80                                                                                                     & 1                                                               & 100                                                            & Yes (0.5) & 96.92                                                                             \\\hline
15                      & 88                                                                                                     & 1                                                               & 100                                                            & Yes (0.5) & 96.36                                              \\\hline                               
\end{tabular}}
\end{table}

In Table \ref{tab3}, we compare the performance of our proposed Derm2Vec method along with the DNN with some of the other conventional machine learning methods used in the literature for the diagnosis of ESD (see Sections \ref{lit} and \ref{others}). We compare Derm2Vec and DNN with other methods such as  Extreme Gradient Boosting (XGBoost), Random Forest (RF), Decision Tree (DT), Na{\"i}ve Bayes (NB), Artificial Neural Network (ANN), K-nearest neighbors (KNN) and Support Vector Classification (SVC). For ANN, we use a simple architecture comprising one hidden layer with two hidden nodes. As far as the choice of kernel for the SVC method is concerned, we chose the "Radial basis function (RBF)" \citep{svc:1}. For the KNN method, we use $K=5$. The different hyperparamters that we chose for Random forests are $n\_estimators=100$ and $max\_depth=3$. Similarly, for XGBoost the hyperparameters selected were $learning\_rate=0.05$,  $n\_estimators=300$ and $max\_depth=3$.

\begin{table}[!htp]
\centering
\caption{Comparing Derm2Vec and DNN with other methods such as Extreme Gradient Boosting (XGBoost), Random Forest (RF), Decision Tree (DT), Na{\"i}ve Bayes (NB), Artificial Neural Network (ANN), K-nearest neighbors (KNN) and Support Vector Classification (SVC)}
\label{tab3}
 \scalebox{0.9}{
  \begin{threeparttable}
\begin{tabular}{lc}
\hline
Method        & \begin{tabular}[c]{@{}c@{}}Mean CV \\ score (in \%)\end{tabular} \\ \hline
Derm2Vec      & 96.92                                                            \\\hline   
DNN           & 96.65                                                            \\\hline   
XGBoost       & 95.80                                                            \\\hline   
DT & 93.10                                                            \\\hline   
ANN*          & 74.29                                                            \\\hline   
SVC**           & 82.13                                                            \\\hline   
RF & 51.13                                                            \\\hline   
NB   & 92.68                                                            \\ \hline   
KNN***           & 79.30                           \\ \hline                                
\end{tabular}
\begin{tablenotes}\footnotesize
\item[*]ANN has 1 hidden layer with 2 hidden nodes
\item[**]SVC has RBF kernel
\item[***]K=5 in KNN
\end{tablenotes}
\end{threeparttable}}
\end{table}

The Derm2Vec method is found to be the best performer (when taking the prediction accuracy into account) followed by DNN and Extreme Gradient Boosting as described in Table \ref{tab3}. Both Derm2Vec and DNN perform better than XGBoost, RF, DT, ANN, SVC, NB and KNN.  The mean CV score of Derm2Vec and DNN are $96.92\%$ and $96.65\%$. However, the mean CV score of XGBoost, DT, ANN, SVC, RF, NB and KNN are $95.80\%$, $93.10\%$, $74.29\%$, $82.13\%$, $51.13\%$, $92.68\%$ and $79.30\%$ respectively.

\section{Conclusion} \label{conclusion}
In this paper, we propose a novel hybrid deep learning approach i.e. Derm2Vec for the diagnosis of the Erythemato-Squamous disease (ESD) that to the best of our knowledge, hasn't been reported in the literature. Also, we find that there haven't been many works reported in the literature regarding the applications of deep neural networks for the classification of ESD. Although the literature is replete with works that used conventional machine learning methods (namely, Random forests, artificial neural networks, Extreme Gradient Boosting, K-nearest neighbors, decision trees, support vector machines and Na{\"i}ve Bayes) for the diagnosis of ESD.

We apply both Derm2Vec and a Deep Neural Network (after tuning the hyperparameters) along with other conventional machine learning methods as mentioned above on a real world dermatology dataset. The Derm2Vec method is found to be the best performer when taking the prediction accuracy into account. Thus, we conclude that our proposed hybrid deep learning approach i.e. Derm2Vec is an effective method for the diagnosis of ESD. We feel that our proposed hybrid deep learning method Derm2Vec can be extended with some modifications in other areas of medicine such as diagnosis of liver disease, cancer prediction, prediction of diabetes and more. We plan to work in this direction in the future.

\section*{References}

\bibliography{Derm2Vecbib}

\begin{thebibliography}{58}
\expandafter\ifx\csname natexlab\endcsname\relax\def\natexlab#1{#1}\fi
\providecommand{\url}[1]{\texttt{#1}}
\providecommand{\href}[2]{#2}
\providecommand{\path}[1]{#1}
\providecommand{\DOIprefix}{doi:}
\providecommand{\ArXivprefix}{arXiv:}
\providecommand{\URLprefix}{URL: }
\providecommand{\Pubmedprefix}{pmid:}
\providecommand{\doi}[1]{\href{http://dx.doi.org/#1}{\path{#1}}}
\providecommand{\Pubmed}[1]{\href{pmid:#1}{\path{#1}}}
\providecommand{\bibinfo}[2]{#2}
\ifx\xfnm\relax \def\xfnm[#1]{\unskip,\space#1}\fi
\bibitem[{Elsayad et~al.(2018)Elsayad, Al-Dhaifallah, and Nassef}]{esd:2}
\bibinfo{author}{A.~M. Elsayad}, \bibinfo{author}{M.~Al-Dhaifallah},
  \bibinfo{author}{A.~M. Nassef},
\newblock \bibinfo{title}{{Analysis and Diagnosis of Erythemato-Squamous
  Diseases Using CHAID Decision Trees}},
\newblock in: \bibinfo{booktitle}{15th International Multi-Conference on
  Systems, Signals and Devices (SSD)}, \bibinfo{publisher}{{IEEE}},
  \bibinfo{year}{2018}. \DOIprefix\doi{10.1109/SSD.2018.8570553}.
\bibitem[{Demiroz et~al.(1998)Demiroz, Govenir, and Ilter}]{esd:1}
\bibinfo{author}{G.~Demiroz}, \bibinfo{author}{H.~A. Govenir},
  \bibinfo{author}{N.~Ilter},
\newblock \bibinfo{title}{{Learning Differential Diagnosis of
  Eryhemato-Squamous Diseases using Voting Feature Intervals}},
\newblock \bibinfo{journal}{{Aritificial Intelligence in Medicine}}
  \bibinfo{volume}{13} (\bibinfo{year}{1998}) \bibinfo{pages}{147--165}.
\bibitem[{Menai and Altayash(2014)}]{esd:3}
\bibinfo{author}{M.~E.~B. Menai}, \bibinfo{author}{N.~Altayash},
\newblock \bibinfo{title}{{Differential Diagnosis of Erythemato-Squamous
  Diseases Using Ensemble of Decision Trees}},
\newblock in: \bibinfo{booktitle}{{ Modern Advances in Applied Intelligence }},
  \bibinfo{year}{2014}, pp. \bibinfo{pages}{369--377}.
\bibitem[{Deng and Liu(2018)}]{dl:7}
\bibinfo{author}{L.~Deng}, \bibinfo{author}{Y.~Liu}, \bibinfo{title}{Deep
  Learning in Natural Language Processing}, \bibinfo{edition}{1} ed.,
  \bibinfo{publisher}{Springer}, \bibinfo{address}{Singapore},
  \bibinfo{year}{2018}. \DOIprefix\doi{10.1007/978-981-10-5209-5}.
\bibitem[{Mamoshina et~al.(2016)Mamoshina, Vieira, Putin, and
  Zhavoronkov}]{dl:6}
\bibinfo{author}{P.~Mamoshina}, \bibinfo{author}{A.~Vieira},
  \bibinfo{author}{E.~Putin}, \bibinfo{author}{A.~Zhavoronkov},
\newblock \bibinfo{title}{Applications of deep learning in biomedicine},
\newblock \bibinfo{journal}{Mol. Pharmaceutics} \bibinfo{volume}{13}
  (\bibinfo{year}{2016}) \bibinfo{pages}{1445--1454}.
\bibitem[{Voulodimos et~al.(2018)Voulodimos, Doulamis, Doulamis, and
  Protopapadakis}]{dl:8}
\bibinfo{author}{A.~Voulodimos}, \bibinfo{author}{N.~Doulamis},
  \bibinfo{author}{A.~Doulamis}, \bibinfo{author}{E.~Protopapadakis},
\newblock \bibinfo{title}{Deep learning for computer vision: A brief review},
\newblock \bibinfo{journal}{Computational Intelligence and Neuroscience}
  (\bibinfo{year}{2018}).
\bibitem[{Cruz and Wishart(2006)}]{lit:1}
\bibinfo{author}{J.~Cruz}, \bibinfo{author}{D.~Wishart},
\newblock \bibinfo{title}{Applications of machine learning in cancer prediction
  and prognosis},
\newblock \bibinfo{journal}{Cancer Informat} \bibinfo{volume}{2}
  (\bibinfo{year}{2006}).
\bibitem[{Gray et~al.(2013)Gray, Aljabar, Heckemann, Hammers, and
  Rueckert}]{lit:3}
\bibinfo{author}{K.~R. Gray}, \bibinfo{author}{P.~Aljabar},
  \bibinfo{author}{R.~A. Heckemann}, \bibinfo{author}{A.~Hammers},
  \bibinfo{author}{D.~Rueckert},
\newblock \bibinfo{title}{Random forest-based similarity measures for
  multi-modal classification of alzheimer's disease},
\newblock \bibinfo{journal}{NeuroImage} \bibinfo{volume}{65}
  (\bibinfo{year}{2013}) \bibinfo{pages}{167--175}.
\bibitem[{Putatunda(2018)}]{arxiv:1}
\bibinfo{author}{S.~Putatunda}, \bibinfo{title}{Care2vec: A deep learning
  approach for the classification of self-care problems in physically disabled
  children}, \bibinfo{howpublished}{arXiv:1812.00715 [cs.LG]},
  \bibinfo{year}{2018}.
\bibitem[{Cheng et~al.(2013)Cheng, Liu, Xu, Yin, Wong, Tan, Tao, Cheng, Aung,
  and Wong}]{lit:4}
\bibinfo{author}{J.~Cheng}, \bibinfo{author}{J.~Liu}, \bibinfo{author}{Y.~Xu},
  \bibinfo{author}{F.~Yin}, \bibinfo{author}{D.~W.~K. Wong},
  \bibinfo{author}{N.-M. Tan}, \bibinfo{author}{D.~Tao}, \bibinfo{author}{C.-Y.
  Cheng}, \bibinfo{author}{T.~Aung}, \bibinfo{author}{T.~Y. Wong},
\newblock \bibinfo{title}{Superpixel classification based optic disc and optic
  cup segmentation for glaucoma screening},
\newblock \bibinfo{journal}{IEEE Transactions on Medical Imaging}
  \bibinfo{volume}{32} (\bibinfo{year}{2013}) \bibinfo{pages}{1019--1032}.
\bibitem[{Tang et~al.(2013)Tang, Niemeijer, Reinhardt, Garvin, and
  Abramoff}]{lit:5}
\bibinfo{author}{L.~Tang}, \bibinfo{author}{M.~Niemeijer},
  \bibinfo{author}{J.~M. Reinhardt}, \bibinfo{author}{M.~K. Garvin},
  \bibinfo{author}{M.~D. Abramoff},
\newblock \bibinfo{title}{Splat feature classification with application to
  retinal hemorrhage detection in fundus images},
\newblock \bibinfo{journal}{IEEE Transactions on Medical Imaging}
  \bibinfo{volume}{32} (\bibinfo{year}{2013}) \bibinfo{pages}{364--375}.
\bibitem[{Luo et~al.(2014)Luo, Sohani, Hochberg, and Szolovits}]{lit:6}
\bibinfo{author}{Y.~Luo}, \bibinfo{author}{A.~R. Sohani},
  \bibinfo{author}{E.~P. Hochberg}, \bibinfo{author}{P.~Szolovits},
\newblock \bibinfo{title}{Automatic lymphoma classification with sentence
  subgraph mining from pathology reports},
\newblock \bibinfo{journal}{Journal of the American Medical Informatics
  Association} \bibinfo{volume}{21} (\bibinfo{year}{2014})
  \bibinfo{pages}{824--832}.
\bibitem[{Guvenir and Emeksiz(2000)}]{esd:4}
\bibinfo{author}{H.~Guvenir}, \bibinfo{author}{N.~Emeksiz},
\newblock \bibinfo{title}{An expert system for the differential diagnosis of
  erythemato-squamous diseases},
\newblock \bibinfo{journal}{Expert Systems with Applications}
  \bibinfo{volume}{18} (\bibinfo{year}{2000}) \bibinfo{pages}{43--49}.
\bibitem[{Ubeyli(2009)}]{esd:5}
\bibinfo{author}{E.~D. Ubeyli},
\newblock \bibinfo{title}{Combined neural networks for diagnosis of
  erythemato-squamous diseases},
\newblock \bibinfo{journal}{Expert Systems with Applications}
  \bibinfo{volume}{36} (\bibinfo{year}{2009}) \bibinfo{pages}{5107--5112}.
\bibitem[{Xie and Wang(2011)}]{esd:6}
\bibinfo{author}{J.~Xie}, \bibinfo{author}{C.~Wang},
\newblock \bibinfo{title}{Using support vector machines with a novel hybrid
  feature selection method for diagnosis of erythemato-squamous diseases},
\newblock \bibinfo{journal}{Expert Systems with Applications}
  \bibinfo{volume}{38} (\bibinfo{year}{2011}) \bibinfo{pages}{5809--5815}.
\bibitem[{Nanni(2006)}]{esd:9}
\bibinfo{author}{L.~Nanni},
\newblock \bibinfo{title}{An ensemble of classifiers for the diagnosis of
  erythemato-squamous diseases},
\newblock \bibinfo{journal}{Neurocomputing} \bibinfo{volume}{69}
  (\bibinfo{year}{2006}) \bibinfo{pages}{842--8845}.
\bibitem[{Menai(2015)}]{esd:11}
\bibinfo{author}{M.~E.~B. Menai},
\newblock \bibinfo{title}{Random forests for automatic differential diagnosis
  of erythemato--squamous diseases},
\newblock \bibinfo{journal}{International Journal of Medical Engineering and
  Informatics} \bibinfo{volume}{7} (\bibinfo{year}{2015}).
\bibitem[{Lekkas and Mikhailov(2010)}]{esd:8}
\bibinfo{author}{S.~Lekkas}, \bibinfo{author}{L.~Mikhailov},
\newblock \bibinfo{title}{Evolving fuzzy medical diagnosis of pima indians
  diabetes and of dermatologica diseases},
\newblock \bibinfo{journal}{Artificial Intelligence in Medicine}
  \bibinfo{volume}{50} (\bibinfo{year}{2010}) \bibinfo{pages}{117--126}.
\bibitem[{Ubeyli and Guler(2005)}]{data:1}
\bibinfo{author}{E.~Ubeyli}, \bibinfo{author}{I.~Guler},
\newblock \bibinfo{title}{Automatic detection of erythemato-squamous diseases
  using adaptive neuro-fuzzy inference systems},
\newblock \bibinfo{journal}{Comput. Biol. Med.} \bibinfo{volume}{35}
  (\bibinfo{year}{2005}) \bibinfo{pages}{421--433}.
\bibitem[{Ubeyli and Dogdu(2010)}]{esd:7}
\bibinfo{author}{E.~D. Ubeyli}, \bibinfo{author}{E.~Dogdu},
\newblock \bibinfo{title}{Automatic detection of erythemato-squamous diseases
  using k-means clustering},
\newblock \bibinfo{journal}{Journal of Medical Systems} \bibinfo{volume}{34}
  (\bibinfo{year}{2010}) \bibinfo{pages}{179--184}.
\bibitem[{Badrinath et~al.(2013)Badrinath, Gopinath, and Ravichandran}]{esd:12}
\bibinfo{author}{N.~Badrinath}, \bibinfo{author}{G.~Gopinath},
  \bibinfo{author}{K.~Ravichandran},
\newblock \bibinfo{title}{Design of automatic detection of erythemato-squamous
  diseases through threshold-based abc-felm algorithm},
\newblock \bibinfo{journal}{Journal of Artificial Intelligence}
  \bibinfo{volume}{6} (\bibinfo{year}{2013}) \bibinfo{pages}{245--256}.
\bibitem[{Bojarczuk et~al.(2001)Bojarczuk, Lopes, and Freitas}]{esd:10}
\bibinfo{author}{C.~C. Bojarczuk}, \bibinfo{author}{H.~S. Lopes},
  \bibinfo{author}{A.~A. Freitas},
\newblock \bibinfo{title}{Data mining with constrained-syntax genetic
  programming: Applications in medical data set},
\newblock in: \bibinfo{booktitle}{Data Analysis in Medicine and Pharmacology
  (IDAMAP- 2001)}, \bibinfo{address}{London, UK}, \bibinfo{year}{2001}.
\bibitem[{Dheeru and Karra~Taniskidou(2017)}]{uci:1}
\bibinfo{author}{D.~Dheeru}, \bibinfo{author}{E.~Karra~Taniskidou},
  \bibinfo{title}{{UCI} machine learning repository},
  \bibinfo{howpublished}{Available:
  \url{https://archive.ics.uci.edu/ml/datasets/SCADI}}, \bibinfo{year}{2017}.
  \bibinfo{note}{{[Dataset]}}.
\bibitem[{McCulloch and Pitts(1943)}]{ann:6}
\bibinfo{author}{W.~S. McCulloch}, \bibinfo{author}{W.~Pitts},
\newblock \bibinfo{title}{A logical calculus of the ideas immanent in nervous
  activity},
\newblock \bibinfo{journal}{The bulletin of mathematical biophysics}
  \bibinfo{volume}{5} (\bibinfo{year}{1943}) \bibinfo{pages}{115--133}.
\bibitem[{Rosenblatt(1962)}]{ann:3}
\bibinfo{author}{F.~Rosenblatt}, \bibinfo{title}{Principles of Neurodynamics:
  Perceptrons and the Theory of Brain Mechanisms}, \bibinfo{publisher}{Spartan
  Books}, \bibinfo{address}{Washington}, \bibinfo{year}{1962}.
\bibitem[{Rumelhart et~al.(1986)Rumelhart, Hinton, and Williams}]{ann:4}
\bibinfo{author}{D.~E. Rumelhart}, \bibinfo{author}{G.~E. Hinton},
  \bibinfo{author}{R.~J. Williams},
\newblock \bibinfo{title}{Learning internal representations by error
  propagation},
\newblock in: \bibinfo{editor}{D.~E. Rumelhart}, \bibinfo{editor}{J.~L.
  McClelland}, \bibinfo{editor}{C.~PDP Research~Group} (Eds.),
  \bibinfo{booktitle}{Parallel Distributed Processing: Explorations in the
  Microstructure of Cognition, Vol. 1}, \bibinfo{publisher}{MIT Press},
  \bibinfo{address}{Cambridge, MA, USA}, \bibinfo{year}{1986}, pp.
  \bibinfo{pages}{318--362}.
\bibitem[{Rosenblatt(1958)}]{ann:8}
\bibinfo{author}{F.~Rosenblatt},
\newblock \bibinfo{title}{The perceptron: A probabilistic model for information
  storage and organization in the brain},
\newblock \bibinfo{journal}{Psychological Review} \bibinfo{volume}{65}
  (\bibinfo{year}{1958}) \bibinfo{pages}{386--408}.
\bibitem[{Cheng and Titterington(1994)}]{ann:7}
\bibinfo{author}{B.~Cheng}, \bibinfo{author}{D.~M. Titterington},
\newblock \bibinfo{title}{Neural networks: A review from a statistical
  perspective},
\newblock \bibinfo{journal}{Statistical Science} \bibinfo{volume}{9}
  (\bibinfo{year}{1994}) \bibinfo{pages}{2--30}.
\bibitem[{Bishop(2006)}]{ann:1}
\bibinfo{author}{C.~M. Bishop}, \bibinfo{title}{{Pattern Recognition and
  Machine Learning (Information Science and Statistics)}},
  \bibinfo{publisher}{{Springer-Verlag}}, \bibinfo{address}{{Berlin,
  Heidelberg}}, \bibinfo{year}{2006}.
\bibitem[{Putatunda(2019)}]{sayan:1}
\bibinfo{author}{S.~Putatunda},
\newblock \bibinfo{title}{{Machine Learning: An Introduction}},
\newblock in: \bibinfo{editor}{A.~K. Laha} (Ed.), \bibinfo{booktitle}{Advances
  in Analytics and Applications}, {Springer Proceedings in Business and
  Economics}, \bibinfo{publisher}{{Springer Nature Singapore Pte Ltd.}},
  \bibinfo{year}{2019}, pp. \bibinfo{pages}{1--9}.
  \DOIprefix\doi{https://doi.org/10.1007/978-981-13-1208-3_1}.
\bibitem[{Larsen(1999)}]{ann:2}
\bibinfo{author}{J.~Larsen}, \bibinfo{title}{Introduction to Artificial Neural
  Network}, \bibinfo{edition}{1st} ed., \bibinfo{publisher}{Technical
  University of Denmark}, \bibinfo{year}{1999}.
\bibitem[{Kumar et~al.(2012)Kumar, Singh, Ghosh, and Anand}]{ann:10}
\bibinfo{author}{A.~Kumar}, \bibinfo{author}{M.~Singh},
  \bibinfo{author}{S.~Ghosh}, \bibinfo{author}{A.~Anand},
\newblock \bibinfo{title}{Weather forecasting model using artificial neural
  network},
\newblock \bibinfo{journal}{Procedia Technology} \bibinfo{volume}{4}
  (\bibinfo{year}{2012}) \bibinfo{pages}{311--318}.
\bibitem[{Baxt(1995)}]{ann:11}
\bibinfo{author}{W.~G. Baxt},
\newblock \bibinfo{title}{Application of neural networks to clinical medicine},
\newblock \bibinfo{journal}{Lancet} \bibinfo{volume}{346}
  (\bibinfo{year}{1995}) \bibinfo{pages}{1135--1138}.
\bibitem[{Eng et~al.(2008)Eng, Li, Wang, and Lee}]{ann:12}
\bibinfo{author}{M.~H. Eng}, \bibinfo{author}{Y.~Li}, \bibinfo{author}{Q.-G.
  Wang}, \bibinfo{author}{T.~H. Lee},
\newblock \bibinfo{title}{Forecast forex with ann using fundamental data},
\newblock in: \bibinfo{booktitle}{2008 International Conference on Information
  Management, Innovation Management and Industrial Engineering},
  \bibinfo{publisher}{IEEE}, \bibinfo{address}{Taipei, Taiwan},
  \bibinfo{year}{2008}. \DOIprefix\doi{10.1109/ICIII.2008.302}.
\bibitem[{Laha and Putatunda(2018)}]{sayan:3}
\bibinfo{author}{A.~Laha}, \bibinfo{author}{S.~Putatunda},
\newblock \bibinfo{title}{Real time location prediction with taxi-gps data
  streams},
\newblock \bibinfo{journal}{{Transportation Research Part C: Emerging
  Technologies}} \bibinfo{volume}{92} (\bibinfo{year}{2018})
  \bibinfo{pages}{298--322}.
\bibitem[{Laha and Putatunda(2017)}]{sayan:2}
\bibinfo{author}{A.~K. Laha}, \bibinfo{author}{S.~Putatunda},
\newblock \bibinfo{title}{{Travel Time Prediction for GPS Taxi Data Streams}},
\newblock \bibinfo{journal}{{Indian Institute of Management Ahmedabad, Working
  Paper No. 2017-03-03}}  (\bibinfo{year}{2017}).
\bibitem[{Putatunda(2017)}]{thesis:sayan}
\bibinfo{author}{S.~Putatunda}, \bibinfo{title}{Streaming Data: New Models and
  Methods with Applications in the Transportation Industry}, Ph.D. thesis,
  Indian Institute of Management Ahmedabad, \bibinfo{year}{2017}.
\bibitem[{Bishop(1995)}]{ann:9}
\bibinfo{author}{C.~M. Bishop}, \bibinfo{title}{{Neural Networks for Pattern
  Recognition}}, \bibinfo{number}{{0198538642}}, \bibinfo{publisher}{{Oxford
  University Press, Inc.}}, \bibinfo{address}{{New York, NY, USA}},
  \bibinfo{year}{1995}.
\bibitem[{Goodfellow et~al.(2016)Goodfellow, Bengio, and Courville}]{dl:1}
\bibinfo{author}{I.~Goodfellow}, \bibinfo{author}{Y.~Bengio},
  \bibinfo{author}{A.~Courville}, \bibinfo{title}{Deep Learning},
  \bibinfo{publisher}{MIT Press}, \bibinfo{year}{2016}.
\bibitem[{Chollet(2017)}]{dl:5}
\bibinfo{author}{F.~Chollet}, \bibinfo{title}{{Deep Learning with Python}},
  \bibinfo{edition}{1st} ed., \bibinfo{publisher}{{Manning Publications Co.}},
  \bibinfo{year}{2017}.
\bibitem[{Wang et~al.(2012)Wang, He, and Prokhorov}]{auto:1}
\bibinfo{author}{J.~Wang}, \bibinfo{author}{H.~He}, \bibinfo{author}{D.~V.
  Prokhorov},
\newblock \bibinfo{title}{A folded neural network autoencoder for
  dimensionality reduction},
\newblock \bibinfo{journal}{Procedia Computer Science} \bibinfo{volume}{13}
  (\bibinfo{year}{2012}) \bibinfo{pages}{120--127}.
\bibitem[{Ghosh et~al.(2018)Ghosh, Olewnik, and Lewis}]{dl:9}
\bibinfo{author}{D.~Ghosh}, \bibinfo{author}{A.~Olewnik},
  \bibinfo{author}{K.~Lewis},
\newblock \bibinfo{title}{Application of autoencoders in cyber-empathic
  design},
\newblock \bibinfo{journal}{Design Science} \bibinfo{volume}{4}
  (\bibinfo{year}{2018}).
\bibitem[{Blaschke et~al.(2018)Blaschke, Olivecrona, Engkvist, Bajorath, and
  Chen}]{dl:10}
\bibinfo{author}{T.~Blaschke}, \bibinfo{author}{M.~Olivecrona},
  \bibinfo{author}{O.~Engkvist}, \bibinfo{author}{J.~Bajorath},
  \bibinfo{author}{H.~Chen},
\newblock \bibinfo{title}{Application of generative autoencoder in de novo
  molecular design},
\newblock \bibinfo{journal}{Molecular Informatics} \bibinfo{volume}{37}
  (\bibinfo{year}{2018}).
\bibitem[{Breiman et~al.(1984)Breiman, Friedman, Stone, and Olshen}]{dt:1}
\bibinfo{author}{L.~Breiman}, \bibinfo{author}{J.~Friedman},
  \bibinfo{author}{C.~J. Stone}, \bibinfo{author}{R.~Olshen},
  \bibinfo{title}{Classification and regression trees},
  \bibinfo{publisher}{{Taylor \& Francis}}, \bibinfo{year}{1984}.
\bibitem[{Loh(2014)}]{stat:5}
\bibinfo{author}{W.-Y. Loh},
\newblock \bibinfo{title}{Fifty years of classification and regression trees},
\newblock \bibinfo{journal}{International Statistical Review}
  \bibinfo{volume}{82} (\bibinfo{year}{2014}) \bibinfo{pages}{329--348}.
\bibitem[{Chen and Guestrin(2016)}]{xgb:1}
\bibinfo{author}{T.~Chen}, \bibinfo{author}{C.~Guestrin},
\newblock \bibinfo{title}{{{XGBoost}: A Scalable Tree Boosting System}},
\newblock in: \bibinfo{booktitle}{{Proceedings of the 22nd ACM SIGKDD
  International Conference on Knowledge Discovery and Data Mining}}, {KDD '16},
  \bibinfo{publisher}{{ACM}}, \bibinfo{address}{{New York, NY, USA}},
  \bibinfo{year}{2016}, pp. \bibinfo{pages}{{785--794}}.
\bibitem[{Breiman(2001)}]{stat:6}
\bibinfo{author}{L.~Breiman},
\newblock \bibinfo{title}{Random forests},
\newblock \bibinfo{journal}{Machine Learning} \bibinfo{volume}{45}
  (\bibinfo{year}{2001}) \bibinfo{pages}{5--32}. \bibinfo{note}{Kluwer Academic
  Publishers, Hingham, MA, USA}.
\bibitem[{Cover and Hart(1967)}]{knn:1}
\bibinfo{author}{T.~Cover}, \bibinfo{author}{P.~Hart},
\newblock \bibinfo{title}{Nearest neighbor pattern classification},
\newblock \bibinfo{journal}{{IEEE Trans. Inf. Theor.}} \bibinfo{volume}{13}
  (\bibinfo{year}{1967}) \bibinfo{pages}{21--27}.
\bibitem[{Cortes and Vapnik(1995)}]{svc:1}
\bibinfo{author}{C.~Cortes}, \bibinfo{author}{V.~Vapnik},
\newblock \bibinfo{title}{{Support-Vector Networks}},
\newblock \bibinfo{journal}{{Mach. Learn.}} \bibinfo{volume}{20}
  (\bibinfo{year}{1995}) \bibinfo{pages}{{273--297}}.
\bibitem[{Webb(2010)}]{nb:1}
\bibinfo{author}{G.~I. Webb}, \bibinfo{title}{{Na{\"\i}ve Bayes}},
  \bibinfo{publisher}{Springer}, \bibinfo{address}{Boston, MA},
  \bibinfo{year}{2010}, pp. \bibinfo{pages}{713--714}.
\bibitem[{Hastie et~al.(2009)Hastie, Tibshirani, and Friedman}]{stat:1}
\bibinfo{author}{T.~Hastie}, \bibinfo{author}{R.~Tibshirani},
  \bibinfo{author}{J.~Friedman}, \bibinfo{title}{The Elements of Statistical
  Learning}, Springer series in statistics, \bibinfo{edition}{2nd} ed.,
  \bibinfo{publisher}{Springer}, \bibinfo{year}{2009}. \bibinfo{note}{New
  York}.
\bibitem[{James et~al.(2013)James, Witten, Hastie, and Tibshirani}]{stat:2}
\bibinfo{author}{G.~James}, \bibinfo{author}{D.~Witten},
  \bibinfo{author}{T.~Hastie}, \bibinfo{author}{R.~Tibshirani},
  \bibinfo{title}{An Introduction to Statistical Learning with Applications in
  R}, \bibinfo{publisher}{Springer}, \bibinfo{year}{2013}.
\bibitem[{Rossum(1995)}]{py:1}
\bibinfo{author}{G.~Rossum}, \bibinfo{title}{Python Reference Manual},
  \bibinfo{type}{Technical Report}, {CWI (Centre for Mathematics and Computer
  Science)}, \bibinfo{address}{{Amsterdam, The Netherlands, The Netherlands}},
  \bibinfo{year}{1995}.
\bibitem[{Pedregosa et~al.(2011)Pedregosa, Varoquaux, Gramfort, Michel,
  Thirion, Grisel, Blondel, Prettenhofer, Weiss, Dubourg, Vanderplas, Passos,
  Cournapeau, Brucher, Perrot, and Duchesnay}]{sci:1}
\bibinfo{author}{F.~Pedregosa}, \bibinfo{author}{G.~Varoquaux},
  \bibinfo{author}{A.~Gramfort}, \bibinfo{author}{V.~Michel},
  \bibinfo{author}{B.~Thirion}, \bibinfo{author}{O.~Grisel},
  \bibinfo{author}{M.~Blondel}, \bibinfo{author}{P.~Prettenhofer},
  \bibinfo{author}{R.~Weiss}, \bibinfo{author}{V.~Dubourg},
  \bibinfo{author}{J.~Vanderplas}, \bibinfo{author}{A.~Passos},
  \bibinfo{author}{D.~Cournapeau}, \bibinfo{author}{M.~Brucher},
  \bibinfo{author}{M.~Perrot}, \bibinfo{author}{E.~Duchesnay},
\newblock \bibinfo{title}{{Scikit-learn: Machine Learning in {P}ython}},
\newblock \bibinfo{journal}{Journal of Machine Learning Research}
  \bibinfo{volume}{12} (\bibinfo{year}{2011}) \bibinfo{pages}{2825--2830}.
\bibitem[{Abadi et~al.(2016)Abadi, Barham, Chen, Chen, Davis, Dean, Devin,
  Ghemawat, Irving, Isard, Kudlur, Levenberg, Monga, Moore, Murray, Steiner,
  Tucker, Vasudevan, Warden, Wicke, Yu, and Zheng}]{py:3}
\bibinfo{author}{M.~Abadi}, \bibinfo{author}{P.~Barham},
  \bibinfo{author}{J.~Chen}, \bibinfo{author}{Z.~Chen},
  \bibinfo{author}{A.~Davis}, \bibinfo{author}{J.~Dean},
  \bibinfo{author}{M.~Devin}, \bibinfo{author}{S.~Ghemawat},
  \bibinfo{author}{G.~Irving}, \bibinfo{author}{M.~Isard},
  \bibinfo{author}{M.~Kudlur}, \bibinfo{author}{J.~Levenberg},
  \bibinfo{author}{R.~Monga}, \bibinfo{author}{S.~Moore},
  \bibinfo{author}{D.~G. Murray}, \bibinfo{author}{B.~Steiner},
  \bibinfo{author}{P.~Tucker}, \bibinfo{author}{V.~Vasudevan},
  \bibinfo{author}{P.~Warden}, \bibinfo{author}{M.~Wicke},
  \bibinfo{author}{Y.~Yu}, \bibinfo{author}{X.~Zheng},
\newblock \bibinfo{title}{Tensorflow: A system for large-scale machine
  learning},
\newblock in: \bibinfo{booktitle}{Proceedings of the 12th USENIX Conference on
  Operating Systems Design and Implementation}, \bibinfo{publisher}{USENIX
  Association}, \bibinfo{address}{Berkeley, CA, USA}, \bibinfo{year}{2016}, pp.
  \bibinfo{pages}{265--283}.
\bibitem[{{Chollet et al.}(2015)}]{py:2}
\bibinfo{author}{{Chollet et al.}}, \bibinfo{title}{Keras},
  \bibinfo{howpublished}{{\url{https://keras.io}}}, \bibinfo{year}{2015}.
\bibitem[{Refaeilzadeh et~al.(2009)Refaeilzadeh, Tang, and Liu}]{eval:1}
\bibinfo{author}{P.~Refaeilzadeh}, \bibinfo{author}{L.~Tang},
  \bibinfo{author}{H.~Liu},
\newblock \bibinfo{title}{Cross-validation},
\newblock in: \bibinfo{editor}{L.~Liu}, \bibinfo{editor}{M.~{\"O}zsu} (Eds.),
  \bibinfo{booktitle}{{Encyclopedia of Database Systems}},
  \bibinfo{publisher}{{Springer, Boston, MA}}, \bibinfo{year}{2009}.
\bibitem[{Srivastava et~al.(2014)Srivastava, Hinton, Krizhevsky, Sutskever, and
  Salakhutdinov}]{dl:11}
\bibinfo{author}{N.~Srivastava}, \bibinfo{author}{G.~Hinton},
  \bibinfo{author}{A.~Krizhevsky}, \bibinfo{author}{I.~Sutskever},
  \bibinfo{author}{R.~Salakhutdinov},
\newblock \bibinfo{title}{{Dropout: A Simple Way to Prevent Neural Networks
  from Overfitting}},
\newblock \bibinfo{journal}{Journal of Machine Learning Research}
  \bibinfo{volume}{15} (\bibinfo{year}{2014}) \bibinfo{pages}{1929--1958}.

\end{thebibliography}

\end{document}